\documentclass[letterpaper, 10 pt, conference]{style/ieeeconf}  
\IEEEoverridecommandlockouts{}                              

\usepackage[dvipsnames]{xcolor} 
\usepackage[normalem]{ulem} 
\newcommand{\stkout}[1]{\ifmmode\text{\sout{\ensuremath{#1}}}\else\sout{#1}\fi}

\usepackage{graphicx} 
\usepackage{amsmath} 
\usepackage{amssymb}  
\usepackage{physics} 
\usepackage[font=small]{caption} 
\usepackage{subcaption} 
\usepackage{siunitx} 
\usepackage{booktabs}
\usepackage{acro} 
\usepackage{multirow} 
\usepackage[normalem]{ulem} 

\usepackage{tikz} 
\usepackage{adjustbox} 
\usetikzlibrary{decorations.pathreplacing, positioning, calc, backgrounds, arrows.meta}

\makeatletter
\let\NAT@parse\undefined{}
\makeatother
\usepackage{hyperref} 

\newcommand*{\T}{^\top}

\newcommand{\rte}[1]{RTE\(_{#1}\)}

\newcommand{\Mc}{\mathcal{M}}
\newcommand{\Nc}{\mathcal{N}}

\newcommand{\Pc}{\mathcal{P}}

\newcommand{\Xc}{\mathcal{X}}

\DeclareAcronym{ate}{short=ATE, long=Absolute Trajectory Error}
\DeclareAcronym{rte}{short=RTE, long=Relative Trajectory Error}
\DeclareAcronym{wrte}{short=RTE\(_j\), long=windowed Relative Trajectory Error}
\DeclareAcronym{icp}{short=ICP, long=Iterative Closest Point}
\DeclareAcronym{imu}{short=IMU, long=Inertial Measurement Unit}
\DeclareAcronym{lidar}{short=LiDAR, long=Light Detection and Ranging}
\DeclareAcronym{lio}{short=LIO, long=\ac{lidar}-Inertial Odometry}
\DeclareAcronym{lo}{short=LO, long=\ac{lidar} Odometry}
\DeclareAcronym{slam}{short=SLAM, long=Simultaneous Localization and Mapping}
\DeclareAcronym{atv}{short=ATV, long=All-Terrain Vehicle}
\DeclareAcronym{pca}{short=PCA, long=Principal Component Analysis}
\DeclareAcronym{sota}{short=SOTA, long=state-of-the-art}

\DeclareAcronym{ours}{short=FORM, long=\textbf{F}ixed-Lag \textbf{O}dometry with \textbf{R}eparative \textbf{M}apping}
\DeclareAcronym{ours_plain}{short=FORM, long=Fixed-Lag Odometry with Reparative Mapping}

\DeclareAcronym{n20}{short=N20, long=Newer Stereo-Cam}
\DeclareAcronym{n21}{short=N21, long=Newer Multi-Cam}
\DeclareAcronym{h22}{short=H22, long=Hilti 2022}
\DeclareAcronym{os}{short=OS, long=Oxford Spires}
\DeclareAcronym{mc}{short=MC, long=Multi-Campus}
\DeclareAcronym{cu}{short=CU, long=CU-Multi}
\DeclareAcronym{bg}{short=BG, long=Botanic Gardens}

\definecolor{color0}{HTML}{0173b2} 
\definecolor{color1}{HTML}{de8f05} 
\definecolor{color2}{HTML}{029e73} 
\definecolor{color3}{HTML}{d55e00}
\definecolor{color4}{HTML}{cc78bc}
\definecolor{color5}{HTML}{ca9161}
\definecolor{color6}{HTML}{fbafe4}
\definecolor{color7}{HTML}{949494}
\definecolor{color8}{HTML}{ece133}
\definecolor{color9}{HTML}{56b4e9}

\definecolor{slow}{gray}{0.4}

\newcommand{\remove}[1]{\textcolor{BrickRed}{\stkout{#1}}}

\renewcommand{\remove}[1]{}

\title{\acs{ours}\@: \acl{ours_plain} \\ utilizing Rotating \acs{lidar} Sensors}

\author{
Easton R. Potokar, Taylor Pool, Daniel McGann, and Michael Kaess
}

\begin{document}

\maketitle

\begin{abstract}
    \acf{lidar} sensors have become a de-facto sensor for many robot state estimation tasks, spurring development of many \acf{lo} methods in recent years.
    While some smoothing-based \ac{lo} methods have been proposed, most require matching against multiple scans, resulting in sub-real-time performance.
    Due to this, most prior works estimate a single state at a time and are ``submap''-based. This architecture propagates any error in  pose estimation to the fixed submap and can cause jittery trajectories and degrade future registrations.
    We propose \acf{ours}, a \ac{lo} method that performs smoothing over a densely connected factor graph while utilizing a single iterative map for matching. This allows for both real-time performance and active correction of the local map as pose estimates are further refined. We evaluate on a wide variety of datasets to show that \ac{ours} is robust, accurate, real-time, and provides smooth trajectory estimates when compared to prior state-of-the-art \ac{lo} methods.
\end{abstract}

\section{Introduction}\label{sec:intro}

Since their introduction, \acf{lidar} sensors have become a popular sensor for many robot state estimation tasks due to their  precision, long range capabilities, and high sensor rate. They have made their way into applications such as unmanned aerial vehicles, quadrupeds, and autonomous ground vehicles, with significant improvements in state estimation accuracy and mapping fidelity over prior sensor modalities~\cite{leeLiDAROdometrySurvey2024}.

For other modalities such as cameras, most \ac{sota} odometry methods estimate a window of previous poses simultaneously, a process known as smoothing~\cite{huangVisualInertialNavigationConcise2019}. Smoothing enables systems to use new information to refine past states, resulting in more accurate and consistent odometry estimates. Some have applied smoothing to \ac{lo}, but generally attempt to perform \ac{icp} against multiple previous scans individually, which results in an over-representation of the environment and sub-real-time performance. Others have utilized GPU-acceleration to achieve real-time smoothing performance, or are part of a large sensor fusion pipeline, but, to the author's knowledge, there has been no real-time \ac{lidar}-only smoothing odometry methods.

To overcome these challenges, most current \acf{lo} methods adopt designs akin to filtering methods in that they estimate a single pose at a time. To utilize information from multiple previous scans, these approaches aggregate multiple scans into a single ``submap'' or ``local map''~\cite{vizzoKISSICPDefensePointtoPoint2023,leeGenZICPGeneralizableDegeneracyRobust2025}. This provides more context for the \ac{icp} step, but remains real-time due to requiring only a single round of matching. Unfortunately, these submaps inherit any errors that may occur in state estimation and are never repaired. This, along with the submap down-sampling techniques, can propagate error and cause irregular trajectories with significant jumps in pose estimation. These choppy trajectories can lead to inaccurate velocity estimation and can be disastrous for downstream tasks such as planning or control.

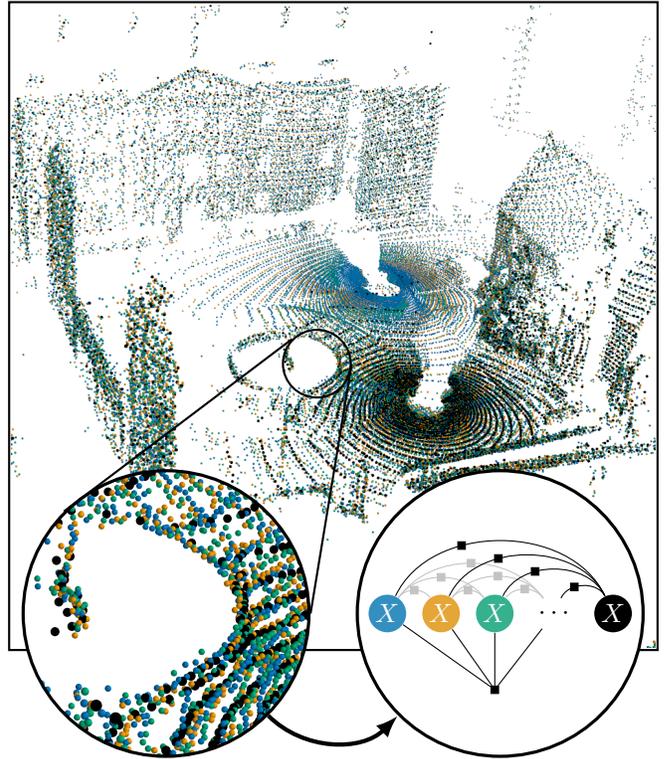
\begin{figure}[t]
    \centering
    \begin{tikzpicture}[
            CIRCLE/.style={circle, draw, very thick, inner sep=0pt},
            FAC/.style={rectangle, minimum size=1mm, inner sep=0pt, draw, fill},
            OLD/.style={fill=color7!55, draw=color7!55},
        ]
        \coordinate (zoom) at (-0.22,-0.5);
        \pgfmathsetlengthmacro{\circleradius}{0.22\columnwidth};
        \pgfmathsetlengthmacro{\circlediameter}{2.0 * \circleradius};

        \node[inner sep=1pt, draw, thick] (image) at (0,0) {
            \includegraphics[width=0.99\columnwidth]{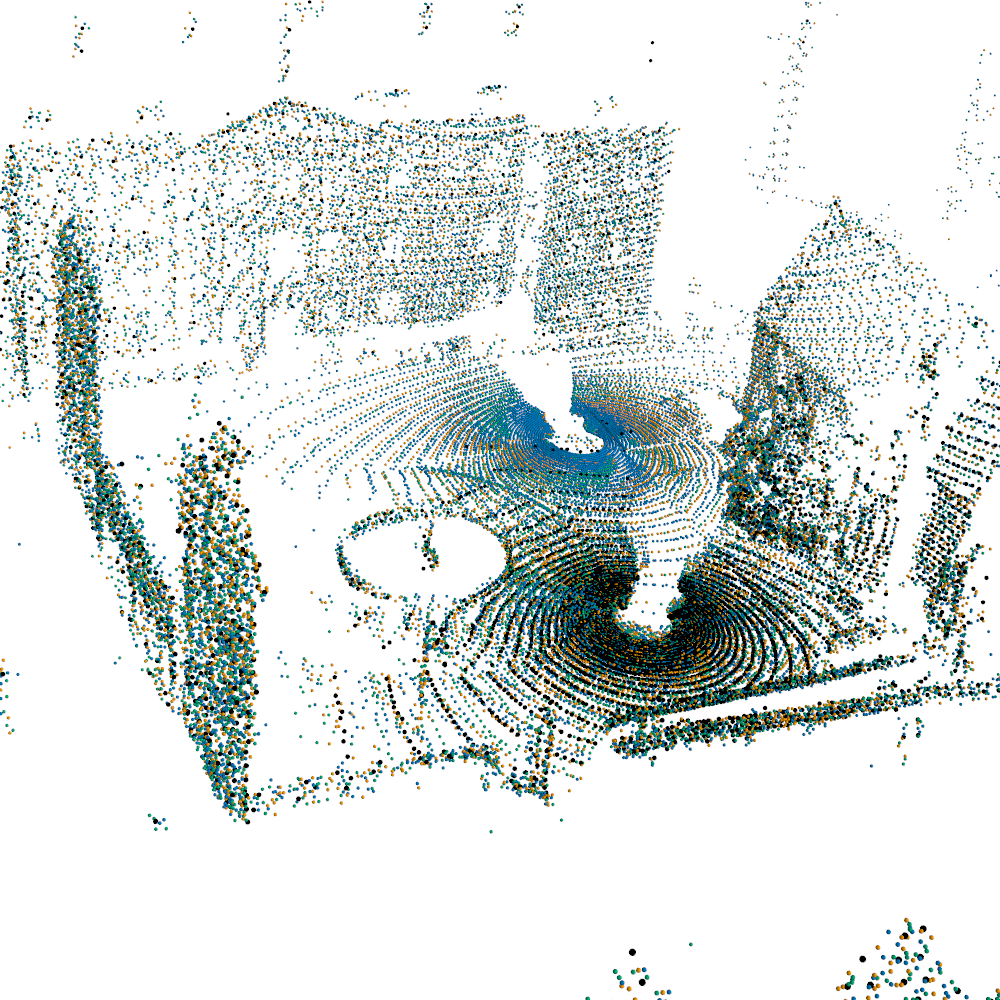}
        };

        \begin{scope}[shift={(image.south west)}, xshift=21mm, yshift=5mm]
            \begin{scope}
                \path[clip] (0,0) circle [x radius=\circleradius, y radius=\circleradius];
                \node[inner sep=0pt] {
                    \includegraphics[width=\circlediameter]{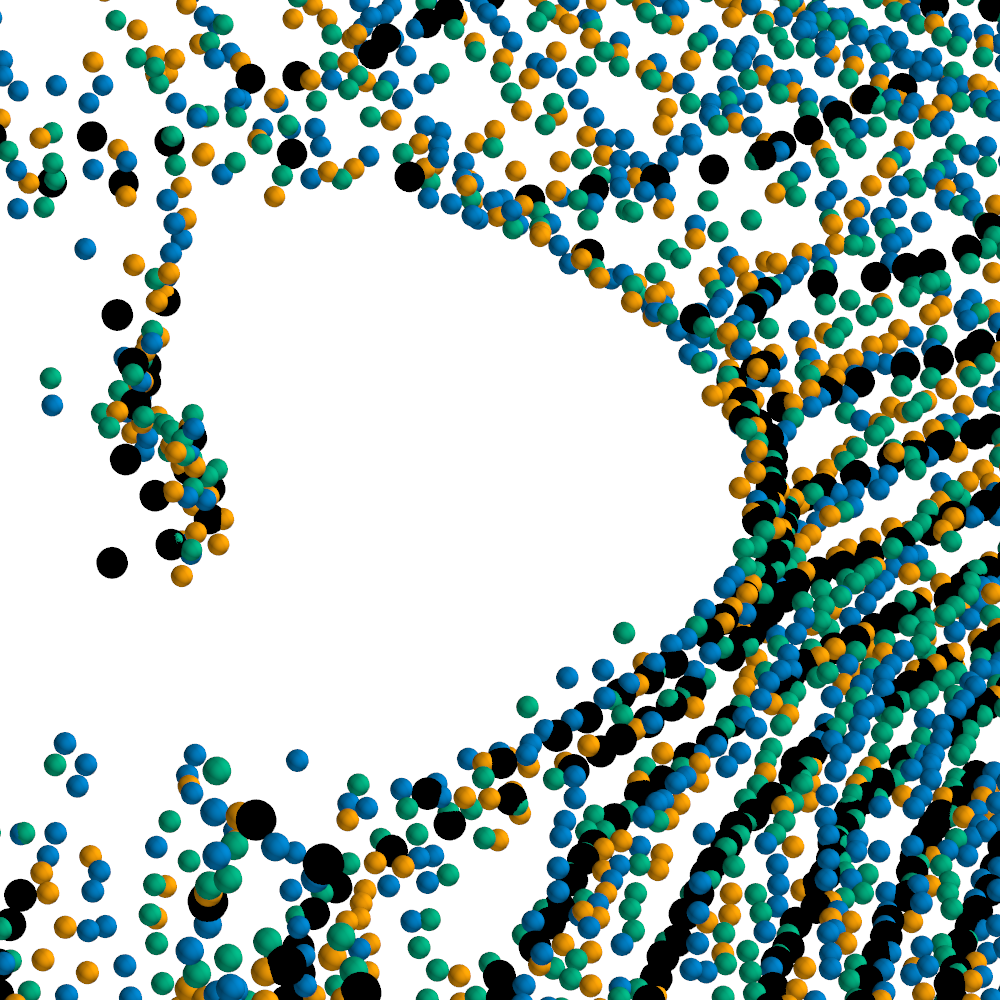}
                };
            \end{scope}
            \node[CIRCLE, minimum size=\circlediameter] (zoom_big) {};
        \end{scope}

        \node[circle, draw, thick, minimum size=9mm, inner sep=0pt] (zoom_small) at (zoom) {};
        \path[draw, thick] (zoom_big.north west) -- (zoom_small.north west);
        \path[draw, thick] (zoom_big.east) -- (zoom_small.east);

        \node[CIRCLE, minimum size=\circlediameter, fill=white] (graph) [right=6mm of zoom_big] {};

        \begin{scope}[node distance=2mm]
            \node[CIRCLE, thin, minimum size=5mm, draw=none, text=white, fill=color0!80] (x_0) at (graph) [xshift=-15mm] {\(X\)};
            \node[CIRCLE, thin, minimum size=5mm, draw=none, text=white, fill=color1!80] (x_1) [right=of x_0] {\(X\)};
            \node[CIRCLE, thin, minimum size=5mm, draw=none, text=white, fill=color2!80] (x_2) [right=of x_1] {\(X\)};

            \node (dots)   [right=of x_2]    {\(\cdots\)};
            \node[CIRCLE, thin, minimum size=5mm, draw=none, text=white, fill=black] (x_3) [right=of dots] {\(X\)};
        \end{scope}

        \node[FAC] (marg) [below=7mm of x_2] {};
        \draw (x_0) -- (marg);
        \draw (x_1) -- (marg);
        \draw (x_2) -- (marg);
        \draw (dots) -- (marg);

        \path[OLD] (x_0) edge[bend left = 55] node[FAC, OLD, pos=0.5] {} (x_1);
        \path[OLD] (x_0) edge[bend left = 55] node[FAC, OLD, pos=0.5] {} (x_2);
        \path[OLD] (x_0) edge[bend left = 55] node[FAC, OLD, pos=0.5] {} (dots);
        \path[OLD] (x_1) edge[bend left = 55] node[FAC, OLD, pos=0.5] {} (x_2);
        \path[OLD] (x_1) edge[bend left = 55] node[FAC, OLD, pos=0.5] {} (dots);
        \path[OLD] (x_2) edge[bend left = 55] node[FAC, OLD, pos=0.5] {} (dots);

        \path (x_0) edge[bend left=65] node[FAC, pos=0.35] {} (x_3);
        \path (x_1) edge[bend left=65] node[FAC, pos=0.35] {} (x_3);
        \path (x_2) edge[bend left=65] node[FAC, pos=0.35] {} (x_3);
        \path (dots) edge[bend left=65] node[FAC, pos=0.35] {} (x_3);

        \path[->, -{Latex[length=3mm]}, ultra thick] (zoom_big.south east) edge[bend right=45] (graph.south west);

    \end{tikzpicture}

    \caption{An example map built using \ac{ours} from the Oxford Spires~\cite{taoOxfordSpiresDataset2024} dataset. Point colors represent which scan the point originated from with black points being from the current scan. The point matches are added as factors to a dense factor graph over a window of prior poses and are optimized in a dense optimization. At each timestep, the map is regenerated with the new, optimized poses. This scheme allows for reparative mapping to occur and minimizes overall pose error.}\label{fig:header}
    \vspace{-1em}
\end{figure}

\newcommand{\fullref}[1]{\nameref*{#1} (Sec.~\ref*{#1})}
\begin{figure*}[t]
    \centering
    \begin{tikzpicture}[
            node distance = -3mm and 3mm,
            font = \small,
            BIGBOX/.style={rectangle, draw=color7!80, fill=color7!20, thick, rounded corners=3pt, minimum height=29mm, minimum width=5.6cm},
            BOXBASE/.style={rectangle, thick, rounded corners=2pt, minimum height=10mm, minimum width=1.5cm},
            BOX1/.style={BOXBASE, draw=color0!60, fill=color0!5},
            BOX2/.style={BOXBASE, draw=color1!60, fill=color1!5},
            BOX3/.style={BOXBASE, draw=color2!60, fill=color2!5},
        ]
        \node (origin) {};

        \node (scan) [align=center] {\ac{lidar} Scan};
        \node[BOX1] (curvature) [above=11mm of scan, align=center] {Compute \\ Curvature};
        \node[BOX1] (extract) [right=of curvature, align=center] {Extract \\ Features};
        \node[BOX1] (normals) [right=of extract, align=center] {Compute \\ Normals};

        \node[BOX2] (init) [right=0.75cm of normals, align=center] {Initialize};
        \node[BOX2] (match) [above right=of init, align=center] {Match \\ Scan};
        \node[BOX2] (linopt) [below right =of init, align=center] {Semi-Lin. \\ Optim.};
        \node[BOX2] (fullopt) [below right=of match, align=center] {Full \\ Optim.};
        \node[BOX2, draw=none, fill=none] (middle) [right=of init] {};

        \node[BOX3] (keyscan) [right=-3mm and 7mm of fullopt, align=center] {Keyscan \\ Selection};
        \node[BOX3] (marg) [right=of keyscan, align=center] {Marg.};
        \node[BOX3] (map) [right=of marg, align=center] {Repair \\ Map};

        \node[text=white] (output_pose) [below=11mm of marg, align=center] {Pose};
        \node[text=white] (output_map) [below=11mm of map, align=center] {Map};
        \node (output) at ($(output_pose)!0.5!(output_map)$) {Smoothed Poses \& Map};
        \node(corner) at (output-|linopt) {};

        \path[->, thick]
        (scan.north) edge (curvature.south)
        (curvature.east) edge (extract.west)
        (extract.east) edge (normals.west)
        (normals.east) edge (init.west)
        (init.north) edge[bend left=30] (match.west)
        ($ (match.south) +(0.4cm,0cm) $) edge[bend left=15]($ (linopt.north) +(0.4cm,0cm) $)
        ($ (linopt.north) -(0.4cm,0cm) $) edge[bend left=15]($ (match.south) -(0.4cm,0cm) $)
        (linopt.east) edge[bend right=20] (fullopt.south)
        (fullopt.east) edge (keyscan.west)
        (keyscan.east) edge (marg.west)
        (marg.east) edge (map.west)
        (marg.south) edge (output_pose.north)
        (map.south) edge (output_map.north);

        \begin{scope}[on background layer]
            \node[BIGBOX,label=above:{\fullref{sec:feats}}] (big_feature) at (extract) {};
            \node[BIGBOX,label=above:{\fullref{sec:icp}}] (big_match) at (middle) {};
            \node[BIGBOX,label=above:{\fullref{sec:marg}}] (big_marg) at (marg) {};
        \end{scope}

        \draw[rounded corners, ->, thick] (output.west) -- (corner.center) -- (big_match.south);

    \end{tikzpicture}
    \caption{The components that make up the \ac{ours} pipeline as described in Section~\ref{sec:methods}. First, point and planar features are extracted. Next, the pose is initialized, \ac{icp} iterations are performed utilizing a semi-linearized dense optimization over past poses, and a full nonlinear optimization follows. Finally, keyscan management is done, poses are marginalized, and a new map is generated from the smoothed poses.}\label{fig:flow}
\end{figure*}
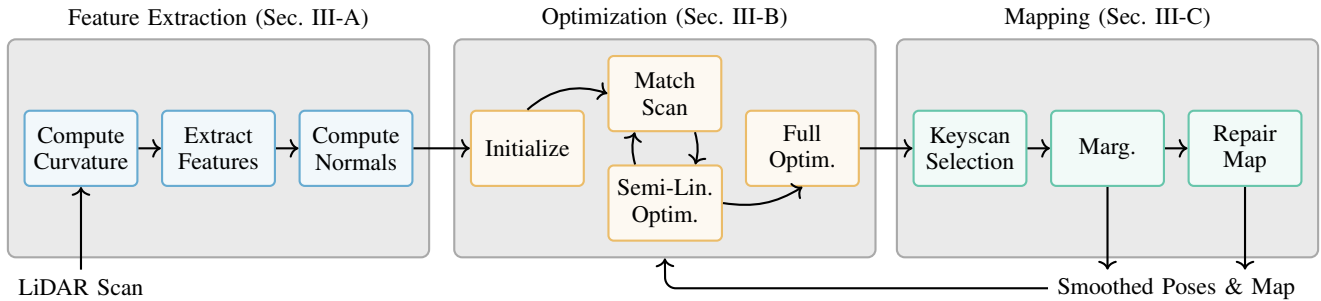

In this work, we propose \acf{ours}, a novel \ac{lo} method that performs fixed-lag smoothing to improve pose estimation accuracy while still maintaining real-time performance. Additionally, the map is iteratively updated with the smoothed poses to prevent errors from propagating. All together, this results in smoother trajectories, increased robustness, and more accurate estimates while maintaining real-time performance. More specifically, our contributions are:
\begin{itemize}
    \item We present a novel real-time \ac{lo} method that optimizes over a fixed window of poses and performs map corrections using the smoothed poses.
    \item To ensure \ac{ours} functions across environments, we develop a novel feature extraction method that utilizes both planar and point features from scans.
    \item We perform ablation studies to demonstrate the various important components of our method, including feature types and smoothing.
    \item We empirically validate our method against other state-of-the-art \ac{lo} methods on a variety of datasets and show that \ac{ours} improves accuracy, reduces irregularities, and increases robustness while still retaining real-time performance.
\end{itemize}

Finally, we release our joint Python-C++ codebase as open-source\footnote{\href{https://github.com/rpl-cmu/form/}{https://github.com/rpl-cmu/form/}}, which includes the \ac{ours} implementation and code to easily recreate all experimental results for our method and all prior works.

\section{Related Works}\label{sec:related_works}

Over the past years, a large amount of \ac{lo} work has be performed with a large variance in techniques. They can be split into roughly two categories: smoothing-based and filtering-based.

Fixed-lag smoothing is a popular technique for state estimation that estimates multiple timesteps of poses jointly and results in more accurate results than filtering-based approaches. Due to the time-complexity of the matching step in \ac{icp}, smoothing approaches for \ac{lo} are more uncommon. Some have simply performed multiple \ac{icp}s that match the current scan against a set of prior scans, with resulting \ac{icp} pose solutions being put into a pose graph at the end~\cite{kurdaLidaronlyOdometryBased2025}. However, this essentially linearizes each scan pair and lacks results that demonstrate real-time performance. Most similar to ours, others~\cite{liangHierarchicalEstimationBasedLiDAR2023} utilize a hierarchical approach that matches against a submap with scan-indexed points and also generates a dense factor graph, but to the author's knowledge, still utilize a fixed submap.

Others have leveraged GPU acceleration to achieve real-time smoothing performance~\cite{koideGLIM3DRangeInertial2024} or down-sampling residuals using coresets~\cite{koideTightlyCoupledRange2025}. Unfortunately, not all vehicles have GPUs and both these methods are utilized solely with IMU fusion.

To maintain real-time performance, most current methods perform filtering, only actively estimating a single pose at any given time. Generally, most of these will perform \ac{icp}~\cite{beslMethodRegistration3D1992} on the current scan against a submap comprised of an aggregation of prior scans. Naturally, there are many variants of this process. Some utilize features extracted from the scans, such as planar or edge features~\cite{zhangLOAMLidarOdometry2014,wangFLOAMFastLiDAR2021,shanLeGOLOAMLightweightGroundOptimized2018}, PCA feature extraction~\cite{panMULLSVersatileLiDAR2021}, or even learned semantic features~\cite{chenSuMaEfficientLiDARbased2019}. Rather than perform feature extraction, others utilize the raw points from the scans~\cite{vizzoKISSICPDefensePointtoPoint2023,leeGenZICPGeneralizableDegeneracyRobust2025}. These result in excellent computational speed and robustness, but often have jagged trajectories due to usage of point-to-point residuals and errors accumulating in the sub-map.

Some filtering-based methods utilize more complex techniques, such as continuous-time estimation over the scan~\cite{dellenbachCTICPRealtimeElastic2022,karimiLoLaSLAMLowLatencyLiDAR2021}, or utilizing novel kd-tree structures to compute normals and matching the normals to previous scans~\cite{ferrariMADICPItAll2024}. While these do result in increased accuracy for many datasets, they often are sub-real-time, require parameter tuning, or still have degradation due to submap aggregation.

Our proposed method achieves the best of both these categories; it has the accuracy and degradation-free results of smoothing while still remaining real-time like prior filtering-based approaches.
\section{\acl{ours_plain} LO}\label{sec:methods}

\ac{ours} is comprised of a number of modular parts, starting with feature extraction, followed by \ac{icp} steps and optimization, and finally keyscan management and mapping. These are summarized in Fig.~\ref{fig:flow} and we cover each in depth here. Note that integer system parameters are denoted with a subscripted \(N\), and thresholds with a subscripted \(\delta \).

\subsection{Feature Extraction}\label{sec:feats}
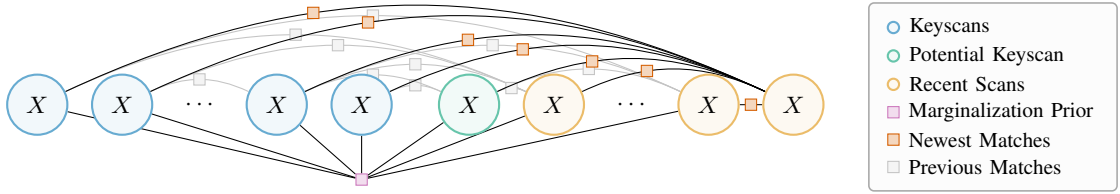
\begin{figure*}[t]
    \centering
    \begin{tikzpicture}[
            node distance = 3mm,
            font = \small,
            VAR/.style={circle, thick, minimum size=8mm},
            VARKEY/.style={circle, thick, minimum size=1.5mm, inner sep=0pt},
            KEY/.style={draw=color0!60, fill=color0!5},
            LAST/.style={draw=color2!60, fill=color2!5},
            RECENT/.style={draw=color1!60, fill=color1!5},
            FAC/.style={rectangle, minimum size=1.5mm, inner sep=0pt},
            MARG/.style={fill=color4!25, draw=color4!100},
            NEW/.style={fill=color3!25, draw=color3!100},
            OLD/.style={fill=color7!10, draw=color7!55},
        ]
        \pgfmathsetmacro{\numkey}{4}
        \pgfmathsetmacro{\numrecent}{3}
        \pgfmathsetmacro{\spacing}{1.5}

        \pgfmathsetmacro{\numkeyhalf}{\numkey / 2 - 1}

        \node[VAR, KEY] (x_k_0)                     {\(X\)};
        \node[VAR, KEY] (x_k_1) [right=of x_k_0]    {\(X\)};
        \node (key_dots)   [right=of x_k_1]    {\(\cdots\)};
        \node[VAR, KEY] (x_k_2) [right=of key_dots] {\(X\)};
        \node[VAR, KEY] (x_k_3) [right=of x_k_2]    {\(X\)};

        \node[VAR, LAST] (x_r_0) [right=6mm of x_k_3] {\(X\)};

        \node[VAR, RECENT] (x_r_1) [right=of x_r_0]         {\(X\)};
        \node (recent_dots)   [right=of x_r_1]       {\(\cdots\)};
        \node[VAR, RECENT] (x_r_2) [right=of recent_dots] {\(X\)};
        \node[VAR, RECENT] (x_r_3) [right=of x_r_2]       {\(X\)};

        \node[FAC, MARG] (marg) [below=5mm of x_k_3] {};
        \begin{scope}[on background layer]
            \draw (x_k_0) -- (marg);
            \draw (x_k_1) -- (marg);
            \draw (x_k_2) -- (marg);
            \draw (x_k_3) -- (marg);
            \draw (x_r_0) -- (marg);
            \draw (x_r_1) -- (marg);
            \draw (x_r_2) -- (marg);
        \end{scope}

        \begin{scope}[on background layer]
            \path[OLD] (x_r_0) edge[bend left=25] node[FAC, OLD, pos=0.5] {} (x_r_2);
            \path[OLD] (x_k_0) edge[bend left=25] node[FAC, OLD, pos=0.5] {} (x_r_2);
            \path[OLD] (x_k_2) edge[bend left=25] node[FAC, OLD, pos=0.5] {} (x_r_2);
            \path[OLD] (x_k_0) edge[bend left=25] node[FAC, OLD, pos=0.5] {} (x_r_1);
            \path[OLD] (x_k_1) edge[bend left=25] node[FAC, OLD, pos=0.5] {} (x_r_1);
            \path[OLD] (x_k_2) edge[bend left=25] node[FAC, OLD, pos=0.5] {} (x_r_1);
            \path[OLD] (x_k_3) edge[bend left=25] node[FAC, OLD, pos=0.5] {} (x_r_0);
            \path[OLD] (x_r_0) edge[bend left=25] node[FAC, OLD, pos=0.5] {} (x_r_1);
            \path[OLD] (x_k_1) edge[bend left=25] node[FAC, OLD, pos=0.5] {} (x_k_2);
            \path[OLD] (x_k_2) edge[bend left=25] node[FAC, OLD, pos=0.5] {} (x_r_0);
        \end{scope}

        \begin{scope}[on background layer]
            \path (x_k_0) edge[bend left=25] node[FAC, NEW, pos=0.35] {} (x_r_3);
            \path (x_k_1) edge[bend left=25] node[FAC, NEW, pos=0.35] {} (x_r_3);
            \path (x_k_2) edge[bend left=25] node[FAC, NEW, pos=0.35] {} (x_r_3);
            \path (x_k_3) edge[bend left=25] node[FAC, NEW, pos=0.35] {} (x_r_3);
            \path (x_r_0) edge[bend left=25] node[FAC, NEW, pos=0.35] {} (x_r_3);
            \path (x_r_1) edge[bend left=25] node[FAC, NEW, pos=0.35] {} (x_r_3);
            \path (x_r_2) edge[bend left=0]  node[FAC, NEW, pos=0.5]  {} (x_r_3);
        \end{scope}

        \node[rectangle, draw=color7!80, fill=color7!3, semithick, rounded corners=2pt, minimum width=33mm, minimum height=25mm] (legend) [above right=-14.5mm and 7mm of x_r_3] {};
        \begin{scope}[font=\footnotesize, node distance=2mm]
            \node[VARKEY, KEY, label=right:Keyscans] (keyscan) [above left=-4mm and -4mm of legend] {};
            \node[VARKEY, LAST, label=right:Potential Keyscan] (oldest) [below=of keyscan] {};
            \node[VARKEY, RECENT, label=right:Recent Scans] (recent) [below=of oldest] {};

            \node[FAC, MARG, label=right:Marginalization Prior] (marg_key) [below=of recent] {};
            \node[FAC, NEW, label=right:Newest Matches] (new_match) [below=of marg_key] {};
            \node[FAC, OLD, label=right:Previous Matches] (old_match) [below=of new_match] {};
        \end{scope}

    \end{tikzpicture}
    \caption{Example of the dense factor graph used in \ac{ours}. Shown are keyscans, recent scans, and the recent scan that is being considered to become a keyscan. During semi-linearized optimization, all the previous matches are linearized to use less compute. The newest matches are also recomputed at every \ac{icp} step.}\label{fig:graph}
\end{figure*}

While the feature extraction used in \ac{ours} is not the main source of novelty, we have found features to have a significant impact on odometry performance and cover it as a demonstration of the joint planar and point feature extraction. We also note that the core components of \ac{ours}, the smoothing optimization and map corrections, are feature invariant and arbitrary features can be utilized.

For our base feature extraction, \ac{ours} utilizes classical LOAM~\cite{zhangLOAMLidarOdometry2014} planar features that are extracted by leveraging the scanline structure of rotating \ac{lidar} sensors. The first step is to mark any points outside of the minimum or maximum range and points along the edge of scanlines as invalid, along with \(N_{\text{neighbor}}\) of their neighbors. We note that since we are not extracting edge features, we do not perform the occlusion or parallel checks that the original LOAM paper used.

Next, an estimated curvature \(\hat{\kappa}_i\) is computed for valid point \(p_i\) utilizing \(N_{\text{neighbor}}\) neighbors in either direction along the scanline~\cite{zhangLOAMLidarOdometry2014},
\begin{align}
    \hat{\kappa}_i & = \Big\lVert\frac{1}{N_{\text{neighbor}}}\sum_{0< j\leq N_{\text{neighbor}}} (p_{i+j} - 2p_i + p_{i-j}) \Big\rVert
\end{align}
Each scanline is split into \(N_{\text{sectors}}\) sectors to ensure features are distributed geometrically, and the \(N_{\text{planar}}\) points with the smallest curvature less than a threshold \(\delta_{\text{planar}} \) in that sector are selected as planar keypoints, making sure to select keypoints that are spaced by at least \(N_{\text{neighbor}}\).

Once features are selected, a normal for each is estimated by solely using the scan. We found utilizing a single scan to be more reliable, does not require re-computation of normals due to the adaptive map described later, and reduces the amount of correlation between poses. Normals are computed by searching adjacent scanlines for the nearest point to the feature \(f\), and selecting \(N_{\text{neighbor}}\) points within a radius \(\delta_{\text{radius}}\) of said nearest point. A normal is then extracted via the smallest eigenvector of the covariance of the selected neighborhood \(\Nc \), making sure \(|\Nc| > 5\),
\begin{align}
    \Sigma_f = \sum_{p\in \Nc} (p - f) (p-f)\T
\end{align}

\begin{figure}[t]
    \setlength{\fboxsep}{0pt}%
    \setlength{\fboxrule}{0.5pt}%
    \fbox{\includegraphics[width=0.98\columnwidth]{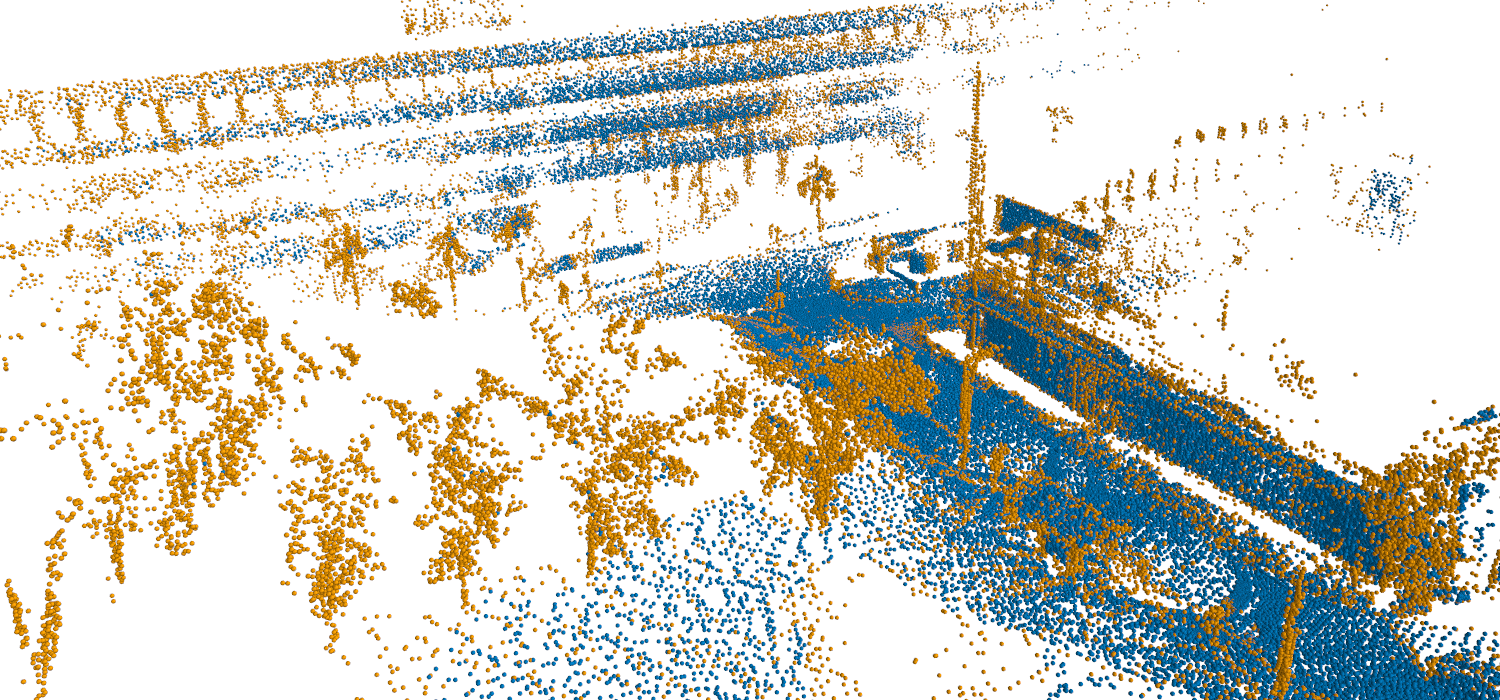}}
    \caption{An example \ac{ours} map, with planar features in blue and point features in orange. Note the trees on the left captured by point features, and walls and ground captured by planar features.}\label{fig:features}
    \vspace{-1em}
\end{figure}

Additionally, we select a number of point features to be utilized. Point features have previously been shown to add robustness when operating in unstructured environments where planar features are less abundant~\cite{potokarComprehensiveEvaluationLiDAR2025}. To ensure our features represent the geometry of the entire scan, we select points features where planar features are not present. We select \(N_{\text{point}}\) point features in each sector in a scanline, making sure they are spaced \(N_{\text{neighbor}}\) points away from other planar or point features and are not points outside the minimum or maximum range. An example map built using these features can be seen in Fig.~\ref{fig:features}. Moving forward, we denote the set of features for scan \(i\) as \(\Pc_i\).

\subsection{Optimization}\label{sec:icp}
Once features are extracted, the optimization stage is performed as shown in the second portion of Fig.~\ref{fig:flow}. First, an initial pose estimate is computed for the scan using a constant velocity model, with velocity extracted using the previous two poses.

This is followed by \ac{icp} iterations, with slight variation to the standard method. We match each keypoint \(p_i \) in \(\Pc_i\) to the closest point \(p_k\) from scan \(k\) in the map \(\Mc \) within threshold distance \(\delta_{\text{match}}\), with all distances computed using the standard Euclidean metric. The generation of \(\Mc \) is described in more detail in the following section. For each match of \(p_i\) and \(p_k\), we add a factor from pose \(X_i\) to \(X_k\), resulting in a densely connected factor graph that can be seen in Fig.~\ref{fig:graph}. The residuals used are the standard point-to-plane metric for planar matches and point-to-point metric for point matches as follows~\cite{rusinkiewiczEfficientVariantsICP2001},
\begin{align}
    r_{\text{planar}}(X_k, X_i) & = (R_k n_k)\T (X_i p_i - X_k p_k) \\
    r_{\text{point}}(X_k, X_i)  & = X_i p_i - X_k p_k
\end{align}

With these new factors, we re-optimize the smoothing window using Levenberg-Marquadt. It should be noted, unlike many factor graph problems such as visual-inertial odometry, the only variables to be estimated are the poses. Since all poses are likely connected, this results in an extremely dense and low variable dimension factor graph. The largest cost of the optimization is actually the high dimensional residual and jacobian computations~\cite{koideGLIM3DRangeInertial2024}.

\newcommand\crule[3][black]{\textcolor{#1}{\rule{#2}{#3}}}

\begin{table*}[t]
    \centering
    \caption{Overview of datasets used in evaluation. The majority of sequences from each dataset are included, with the exception of some with motion beyond what is common in a robot trajectory.}\label{table:datasets}
    \begin{tabular}{*{9}{c}}
        \hline
        Dataset                                                         & Acronym & Year & Setting        & Device       & Lidar          & Beams  & Ground Truth               \\
        \hline
        Newer Stereo-Cam~\cite{ramezaniNewerCollegeDataset2020}         & N20     & 2020 & Outdoor Campus & Handheld     & Ouster OS-1    & 64     & Laser Scanner \& ICP       \\
        Newer Multi-Cam~\cite{zhangMultiCameraLiDARInertial2022}        & N21     & 2021 & Outdoor Campus & Handheld     & Ouster OS-0    & 128    & Laser Scanner \& ICP       \\
        Hilti 2022~\cite{zhangHiltiOxfordDatasetMillimeterAccurate2023} & H22     & 2022 & Indoor Campus  & Handheld     & Hesai PandarXT & 32     & Laser Scanner \& ICP       \\
        Oxford Spires~\cite{taoOxfordSpiresDataset2024}                 & OS      & 2024 & Campus         & Backpack     & Hesai QT64     & 64     & Laser Scanner \& ICP       \\
        Multi-Campus~\cite{nguyenMCDDiverseLargeScale2024}              & MC      & 2024 & Outdoor Campus & Handheld/ATV & Ouster OS-1    & 64/128 & Laser Scanner \& Cont-Opt. \\
        Botanic Garden~\cite{liuBotanicGardenHighQualityDataset2024}    & BG      & 2024 & Trail Road     & ATV          & Velodyne VLP16 & 16     & Laser Scanner \& ICP       \\
        CU-Multi~\cite{albinCUMultiDatasetMultiRobot2025}               & CU      & 2025 & Outdoor Campus & ATV          & Ouster OS-0    & 64     & RTK-GPS Fusion             \\
        \hline
    \end{tabular}
\end{table*}

\begin{table}
    \centering
    \caption{\ac{ours} System Parameters}\label{tab:parameters}
    \begin{tabular}{ccl}
        \multicolumn{3}{c}{Significant Parameters}                                                     \\
        \toprule
        Parameter                   & Value            & Description                                   \\
        \midrule
        \(N_{\text{neighbor}}\)     & 5                & Neighbor points in curvature computation      \\
        \(\delta_{\text{map}}\)     & \SI{0.1}{\meter} & Map insertion threshold                       \\
        \(\delta_{\text{match}}\)   & \SI{0.8}{\meter} & Maximum match distance                        \\
        \(N_{\text{recent}}\)       & 10               & Number of recent scans                        \\
        \(\delta_{\text{key}}\)     & 0.1              & Keyscan criteria threshold                    \\
        \bottomrule                                                                                    \\
        \multicolumn{3}{c}{Insignificant Parameters}                                                   \\
        \toprule
        Parameter                   & Value            & Description                                   \\
        \midrule
        \(N_{\text{sectors}}\)      & 6                & Extraction scanline sector size               \\
        \(N_{\text{point}}\)        & 3                & Point features extracted per scanline sector  \\
        \(N_{\text{planar}}\)       & 50               & Planar features extracted per scanline sector \\
        \(\delta_{\text{planar}} \) & 1.0              & Threshold for planar classification           \\
        \(\delta_{\text{radius}} \) & \SI{1.0}{\meter} & Neighbor radius for normal computation        \\
        \(N_{\text{icp}}\)          & 30               & Number of \ac{icp} iterations                 \\
        \(\delta_{\text{icp}}\)     & 1e-4             & Stopping pose threshold for \ac{icp}          \\
        \(N_{\text{key}}\)          & 50               & Maximum number of keyscans                    \\
        \(N_{\text{marg}}\)         & 10               & Unconnected keyscan marginalization steps     \\
        \bottomrule
    \end{tabular}
\end{table}

Due to this, and since previous poses in the smoothing window will not change significantly, before beginning the \ac{icp} loop we linearize all previous match factors not connected to the current scan. This results in a ``semi-linearized'' optimization that significantly speeds up the \ac{icp} process and allows more time for additional matching iterations to be performed. Also, this optimization is initialized with smoothed pose results from the previous full optimization, not from the previous semi-linearized optimization, in order to avoid poor pose estimates from initial matching that can degrade estimates.

The matching and optimization loop continues until a maximum iteration threshold \(N_{\text{icp}}\) is hit, or until the change in the most recent pose is less than a threshold \(\delta_{\text{icp}}\). Once the \ac{icp} iterations are completed, a full nonlinear optimization is performed without the linearization of previous matches.

\subsection{Mapping}\label{sec:marg}
Once optimization is finalized, keyscanning and map management are performed as in the right of Fig.~\ref{fig:flow}. Keyscanning ensures the map \(\Mc \) continues to cover sufficient geometric area. At any given time, a maximum of \(N_{\text{key}}\) keyscans and \(N_{\text{recent}}\) recent scans are kept. If there are more than \(N_{\text{recent}}\) recent scans, the oldest recent scan \(i\) is considered to become a keyscan based on the metric,
\begin{align}
    \frac{|\text{matches to recent scans}|}{|N_{\text{recent}}| |\Pc_i|} > \delta_{\text{key}}
\end{align}
This roughly measures how important the scan is to the other recent scans, weighted by the number of recent scans and keypoints extracted. If scan \(i\) does not meet this threshold, it is marginalized out of the optimization.

A check is also performed to see if any keyscans are no longer needed. If a keyscan has not been matched to a recent scan in the last \(N_{\text{marg}}\) iterations, it is marginalized out. Additionally, to make sure an unlimited number of keyscans is not possible, the oldest keyscan is marginalized out if the number of keyscans exceeds \(N_{\text{key}}\). In practice we have found this limit is rarely hit.

Next, a new map \(\Mc \) is generated. Any keypoints \(p_i\) from the current scan \(\Pc_i\) with a match distance above a threshold \(\delta_{\text{map}}\) are kept to be inserted into the map. We found adjusting this parameter to provide intuitive results; lower values resulted in a denser map and increased accuracy at a runtime cost, with higher values providing the opposite.

Map points are all stored in their respective scan's local frame. Upon map generation, all of these points are transformed into the world frame using the most recent pose estimates and stored with each point's respective scan index to form the map \(\Mc \). Since the map is generated with smoothed poses, many previous registration errors are removed resulting in a repaired or corrected map. Experiments were done with performing this map generation after each \ac{icp} step, but it was found that initial matches resulted in inaccurate pose estimates and caused degradation in the map, hence the map is only generated at the end of the pipeline.

Finally, we note that \ac{ours} does have a decent number of parameters. However, we have found a large majority of them to have limited impact on performance or are feature extraction related and not a core component of \ac{ours}. The parameters used are all shown along with their subjective significance to performance in Table~\ref{tab:parameters}

\section{Experiment Setup}

We design our experiments with carefully chosen metrics, datasets, and prior works to demonstrate the performance of \ac{ours}. All data-loading and prior work methods were run using the open-source \verb|evalio|~\cite{potokarComprehensiveEvaluationLiDAR2025} library.

\subsection{Datasets}
The list of seven datasets chosen for evaluation can be seen in Table~\ref{table:datasets}, with a total of 64 trajectories. To ensure that our method generalizes well, we have prioritized maximizing the variety of \ac{lidar} brands, \ac{lidar} beam-size, and vehicle motion while selecting only datasets that provide accurate ground truth. All datasets have \ac{lidar} sensor rates of \SI{10}{\hertz}, with the exception of CU-Multi which is \SI{20}{\hertz}.

The trajectories in our chosen datasets range anywhere from around \SI{1}{\minute} to \SI{25}{\minute} long. Due to the number of experiments ran, we limit experiments to running on the first \SI{10}{\minute} of longer trajectories, noting this results in about 6,000 \ac{lidar} scans and is a sufficient sample size for our chosen metrics.

\subsection{Metrics}
\begin{figure}[t]
    \centering
    \includegraphics[width=0.48\textwidth]{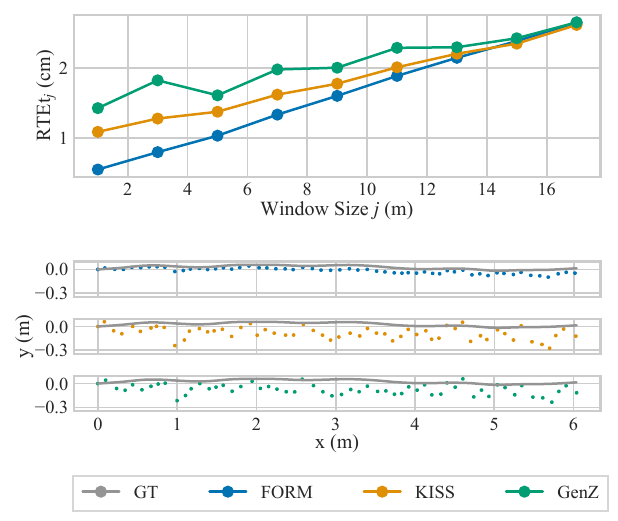}
    \caption{\ac{wrte} small window size demonstration on a trajectory in the \acl{os} dataset. In the above plot, many filtering methods result in higher than expected \ac{wrte} for small windows sizes. This can be seen in the jittery trajectories shown in the other plots, with the exception of \ac{ours}.}\label{fig:window}
    \vspace{-0.5em}
\end{figure}

The most common metric used for evaluating \ac{lo} drift is translational \ac{wrte} with window size \(j\) in meters. To compute this, from the ground truth pose set \(\Xc \), we place all subsequences of metric length \(j\) defined by a change in pose \(\delta \) into the set \(\Xc_j\). We additionally skip any subsequences with an identical ending pose to ensure times when the vehicle is stationary are not overly represented. The metric is then computed by, denoting our trajectory estimate as \(\hat{\Xc}\),
\begin{align}
    \text{RTE}_j(\Xc, \hat{\Xc}) = {\left(\frac{1}{|\Xc_j|} \sum_{\delta_i \in \Xc_j, \hat{\delta}_i \in \hat{\Xc}_j} {\lVert\text{trans}(\delta_{i} \ominus \hat{\delta}_{i})\rVert}^2\right)}^{1/2}
\end{align}
Importantly, varying the window size can provide different context about the results; a smaller window size can capture the local smoothness of a trajectory, while longer window sizes represent longer-term drift quantities. This can be seen in Fig.~\ref{fig:window}, where \ac{wrte} is plotted over window size in meters. Notice the choppy trajectory of prior works, which is well-captured by the small window size \ac{wrte}.

Rotational \ac{wrte} can also be computed, but we found it produces similar trends to translational \ac{wrte} so it is omitted for conciseness.

Finally, we also measure real-time performance by timing the performance of adding a new \ac{lidar} scan to the pipelines, making sure to exclude any time spent on data loading, result saving, etc.

\subsection{Prior Works}
We compare against a number of \ac{sota} \ac{lo} methods that have been open-sourced. This includes the point-to-point submap-based KiSS-ICP~\cite{vizzoKISSICPDefensePointtoPoint2023}, joint point-to-point and point-to-plane submap-based GenZ-ICP~\cite{leeGenZICPGeneralizableDegeneracyRobust2025}, the novel kd-tree matcher MAD-ICP~\cite{ferrariMADICPItAll2024}, and the continuous-time CT-ICP~\cite{dellenbachCTICPRealtimeElastic2022}. Unfortunately, at this time, to the author's knowledge, there is no open-source CPU-based smoothing \ac{lo} methods available for comparison.

All prior works were run with the default parameters found in their open-source repositories. The only exception to this is increasing the number of \ac{icp} iterations and switching the solver to Ceres in CT-ICP\@. Additionally, MAD-ICP is ran with a ``real-time'' flag enabled, resulting in varying performance depending on compute power. While excellent for real-world deployment, it makes benchmarking slightly stochastic. When disabled, MAD-ICP performs at sub-real-time, so we leave it enabled. Finally, moving forward we drop the ``ICP'' at the end of prior works for conciseness. Finally, dewarping was disabled on all prior works except CT-ICP, as it should provide a similar marginal improvement across methods. As mentioned previously, \ac{ours} parameters for all experiments are shown in Table~\ref{tab:parameters}.

\begin{table}
	\centering
	\caption{Ablation study showing the effects of smoothing in \ac{ours}. Values are averaged across dataset sequences, with results in bold representing the best results. Smoothing improves the drift performance of \ac{ours} but has limited impact on the smoothness of the trajectory.}\label{tab:ablation_smoothing}
	\begin{tabular}{c||cccccc}
		\toprule
		D.  & \multicolumn{3}{c}{Filtered} & \multicolumn{3}{c}{Smoothed}                                               \\
		\midrule
		    & \rte{1}                      & \rte{30}                     & Hz   & \rte{1}       & \rte{30}      & Hz   \\
		\midrule
		N20 & \textbf{0.08}                & 0.46                         & 29.6 & \textbf{0.08} & \textbf{0.45} & 24.1 \\
		N21 & \textbf{0.07}                & 0.63                         & 14.1 & \textbf{0.07} & \textbf{0.58} & 12.1 \\
		H22 & 0.14                         & 1.58                         & 40.0 & \textbf{0.13} & \textbf{1.24} & 24.3 \\
		OS  & \textbf{0.08}                & 0.69                         & 30.6 & \textbf{0.08} & \textbf{0.60} & 23.5 \\
		MC  & \textbf{0.09}                & 0.53                         & 21.1 & \textbf{0.09} & \textbf{0.51} & 17.3 \\
		CU  & 0.04                         & 0.31                         & 47.5 & \textbf{0.03} & \textbf{0.20} & 37.3 \\
		BG  & \textbf{0.05}                & 1.00                         & 82.9 & \textbf{0.05} & \textbf{0.82} & 53.0 \\

		\bottomrule
	\end{tabular}
\end{table}

\begin{table}
	\centering
	\caption{Ablation study comparing features used in \ac{ours}. Values are averaged across dataset sequences, with results in bold representing the best results. While the point and planar variant of \ac{ours} does not significantly improve smoothness, it does help with odometry drift at a performance cost.}\label{tab:ablation_feature}
	\begin{tabular}{c||cccccc}
		\toprule
		D.  & \multicolumn{3}{c}{Planar} & \multicolumn{3}{c}{Point \& Planar}                                               \\
		\midrule
		    & \rte{1}                    & \rte{30}                            & Hz   & \rte{1}       & \rte{30}      & Hz   \\
		\midrule
		N20 & 0.09                       & 0.64                                & 43.3 & \textbf{0.08} & \textbf{0.45} & 24.1 \\
		N21 & \textbf{0.07}              & \textbf{0.58}                       & 22.6 & \textbf{0.07} & \textbf{0.58} & 12.1 \\
		H22 & \textbf{0.11}              & \textbf{1.12}                       & 33.6 & 0.13          & 1.24          & 24.3 \\
		OS  & \textbf{0.08}              & 1.01                                & 43.2 & \textbf{0.08} & \textbf{0.60} & 23.5 \\
		MC  & 0.10                       & 0.67                                & 36.7 & \textbf{0.09} & \textbf{0.51} & 17.3 \\
		CU  & 0.04                       & 0.28                                & 64.9 & \textbf{0.03} & \textbf{0.20} & 37.3 \\
		BG  & 0.06                       & 0.87                                & 67.6 & \textbf{0.05} & \textbf{0.82} & 53.0 \\

		\bottomrule
	\end{tabular}
	\vspace{-0.75em}
\end{table}

\section{Experiments}\label{sec:experiments}
We designed a number of experiments to showcase the odometry performance of \ac{ours} and ablations to demonstrate the effectiveness of the various techniques utilized. Across all experiments, we utilize a \rte{1} to represent the smoothness of a trajectory, and \rte{30} for odometry drift. All experiments were run on a 12th-Gen Intel i9 processor.

It should be noted that for \ac{ours}, the pose is saved immediately after being added to the pipeline, not at the time of marginalization. It is likely that \ac{ours} results on marginalized poses would be slightly improved due to the effects of continued smoothing.

\begin{table}
	\centering
	\caption{Average compute speed of the various pipelines in hertz, while excluding trajectories with \rte{30} greater than ten meters. Real-time performance is achieved above the \ac{lidar} sensor rate, \SI{10}{\hertz} for all datasets except CU-Multi which is \SI{20}{\hertz}. \ac{ours} consistently provides real-time performance.}\label{tab:baseline_speed}
	\begin{tabular}{l||ccccc}
		\toprule
		D.  & \acs{ours} & KISS  & GenZ & MAD  & CT   \\
		\midrule
		N20 & 24.1       & 42.3  & 45.9 & 10.6 & 10.2 \\
		N21 & 12.1       & 83.9  & 51.6 & 13.7 & 14.8 \\
		H22 & 24.3       & 224.9 & 67.1 & 40.5 & 15.3 \\
		OS  & 23.5       & 88.3  & 52.8 & 13.8 & 12.9 \\
		MC  & 17.3       & 34.5  & 28.6 & 10.2 & 7.1  \\
		CU  & 37.3       & 92.3  & 49.6 & 21.0 & 11.1 \\
		BG  & 53.0       & 172.9 & 56.5 & 20.3 & 10.8 \\

		\bottomrule
	\end{tabular}
	\vspace{-0.75em}
\end{table}

\begin{table*}
	\centering
	\caption{Both \rte{1} and \rte{30} for all datasets. Strike-throughs indicate a failure to complete, italics sub-real-time performance, \textcolor{color0}{\textbf{bold blue}} the best real-time result, and \textcolor{color0!80}{light blue} the second best. \ac{ours} has competitive drift performance while still maintaining a smooth trajectory and real-time speed.}\label{tab:baseline_main}

	\begin{minipage}{.5\linewidth}
		\centering
		\begin{tabular}{l|c||ccccc}
			\toprule
			\multicolumn{7}{c}{\rte{1}}                                                                                                                                                              \\
			\toprule
			D. & Seq.  & \acs{ours}                        & KISS                        & GenZ                              & MAD                               & CT                                \\
			\midrule
			\multirow[c]{3}{*}{\rotatebox[origin=c]{90}{N20}}
			   & le    & \textcolor{color0}{\textbf{0.04}} & 0.09                        & 0.07                              & \textcolor{color0!80}{0.05}       & 0.07                              \\
			   & pm    & \textcolor{color0}{\textbf{0.14}} & \textcolor{color0!80}{0.18} & \textcolor{slow}{\sout{1.19}}     & 0.19                              & \textcolor{slow}{\textit{1.65}}   \\
			   & se    & \textcolor{color0}{\textbf{0.05}} & 0.17                        & 0.08                              & \textcolor{color0!80}{0.06}       & \textcolor{slow}{\textit{0.07}}   \\
			\midrule
			\multirow[c]{9}{*}{\rotatebox[origin=c]{90}{N21}}
			   & c     & \textcolor{color0}{\textbf{0.04}} & 0.13                        & 0.08                              & \textcolor{color0!80}{0.05}       & 0.05                              \\
			   & me    & \textcolor{color0}{\textbf{0.03}} & 0.05                        & \textcolor{color0!80}{0.04}       & 0.07                              & 0.06                              \\
			   & mh    & \textcolor{color0!80}{0.13}       & 0.21                        & 1.26                              & 1.91                              & \textcolor{color0}{\textbf{0.06}} \\
			   & mm    & \textcolor{color0}{\textbf{0.07}} & 0.10                        & \textcolor{color0!80}{0.10}       & 3.58                              & 0.28                              \\
			   & p     & \textcolor{color0!80}{0.06}       & 0.10                        & 0.08                              & \textcolor{slow}{\sout{9.28}}     & \textcolor{color0}{\textbf{0.05}} \\
			   & qe    & \textcolor{color0!80}{0.05}       & 0.07                        & 0.06                              & 0.09                              & \textcolor{color0}{\textbf{0.04}} \\
			   & qh    & \textcolor{color0!80}{0.11}       & 0.17                        & 0.13                              & 1.15                              & \textcolor{color0}{\textbf{0.08}} \\
			   & qm    & \textcolor{slow}{\textit{0.07}}   & \textcolor{color0!80}{0.12} & \textcolor{color0}{\textbf{0.11}} & 0.16                              & 0.90                              \\
			   & s     & \textcolor{color0!80}{0.05}       & 0.80                        & 0.06                              & \textcolor{color0}{\textbf{0.03}} & 3.88                              \\
			\midrule
			\multirow[c]{6}{*}{\rotatebox[origin=c]{90}{H22}}
			   & atug2 & \textcolor{color0}{\textbf{0.17}} & 1.85                        & 0.64                              & \textcolor{color0!80}{0.32}       & 5.16                              \\
			   & b2    & \textcolor{color0!80}{0.14}       & 0.82                        & 0.14                              & \textcolor{color0}{\textbf{0.11}} & 1.26                              \\
			   & clg2  & \textcolor{color0}{\textbf{0.15}} & 1.55                        & \textcolor{color0!80}{0.20}       & \textcolor{slow}{\sout{0.13}}     & 1.00                              \\
			   & cul1  & \textcolor{color0}{\textbf{0.10}} & 0.64                        & \textcolor{color0!80}{0.62}       & 1.23                              & 4.39                              \\
			   & cul2  & \textcolor{color0}{\textbf{0.08}} & 0.90                        & \textcolor{color0!80}{0.69}       & 0.78                              & 1.09                              \\
			   & cul3  & \textcolor{color0}{\textbf{0.16}} & 4.96                        & \textcolor{slow}{\sout{4.54}}     & \textcolor{color0!80}{1.32}       & 4.39                              \\
			\midrule
			\multirow[c]{13}{*}{\rotatebox[origin=c]{90}{OS}}
			   & bl02  & \textcolor{color0!80}{0.10}       & 0.21                        & 0.35                              & 0.10                              & \textcolor{color0}{\textbf{0.04}} \\
			   & bp01  & \textcolor{color0}{\textbf{0.12}} & 0.38                        & 0.60                              & \textcolor{color0!80}{0.17}       & 0.52                              \\
			   & bp02  & \textcolor{color0!80}{0.05}       & 0.11                        & 0.14                              & 0.06                              & \textcolor{color0}{\textbf{0.05}} \\
			   & bp05  & \textcolor{color0}{\textbf{0.13}} & 0.61                        & 1.27                              & \textcolor{color0!80}{0.16}       & 1.45                              \\
			   & cc01  & \textcolor{color0!80}{0.07}       & 0.21                        & 0.13                              & 0.08                              & \textcolor{color0}{\textbf{0.03}} \\
			   & cc02  & \textcolor{color0!80}{0.05}       & 0.17                        & 0.13                              & 0.07                              & \textcolor{color0}{\textbf{0.04}} \\
			   & cc03  & \textcolor{color0!80}{0.05}       & 0.10                        & 0.08                              & 0.06                              & \textcolor{color0}{\textbf{0.03}} \\
			   & cc05  & \textcolor{color0!80}{0.05}       & 0.14                        & 0.09                              & 0.06                              & \textcolor{color0}{\textbf{0.05}} \\
			   & kc02  & \textcolor{color0}{\textbf{0.08}} & 0.19                        & 0.12                              & \textcolor{color0!80}{0.09}       & 0.22                              \\
			   & kc03  & \textcolor{color0}{\textbf{0.07}} & 0.39                        & 0.21                              & \textcolor{color0!80}{0.09}       & 0.14                              \\
			   & kc04  & \textcolor{color0}{\textbf{0.08}} & 0.59                        & 0.72                              & \textcolor{color0!80}{0.10}       & \textcolor{slow}{\textit{0.04}}   \\
			   & oq01  & \textcolor{color0!80}{0.07}       & 0.13                        & 0.10                              & 0.07                              & \textcolor{color0}{\textbf{0.04}} \\
			   & oq02  & \textcolor{color0}{\textbf{0.07}} & 0.14                        & 0.20                              & \textcolor{color0!80}{0.08}       & \textcolor{slow}{\textit{0.29}}   \\
			\midrule
			\multirow[c]{18}{*}{\rotatebox[origin=c]{90}{MC}}
			   & kd06  & \textcolor{color0}{\textbf{0.10}} & 0.15                        & \textcolor{color0!80}{0.14}       & 0.15                              & \textcolor{slow}{\textit{0.05}}   \\
			   & kd09  & \textcolor{color0}{\textbf{0.13}} & 0.20                        & \textcolor{slow}{\sout{0.19}}     & \textcolor{color0!80}{0.15}       & \textcolor{slow}{\textit{0.06}}   \\
			   & kd10  & \textcolor{color0}{\textbf{0.10}} & \textcolor{color0!80}{0.18} & \textcolor{slow}{\sout{0.23}}     & 0.19                              & \textcolor{slow}{\textit{0.37}}   \\
			   & kn01  & \textcolor{color0}{\textbf{0.10}} & 0.15                        & \textcolor{color0!80}{0.14}       & 0.15                              & \textcolor{slow}{\textit{0.05}}   \\
			   & kn04  & \textcolor{color0}{\textbf{0.10}} & \textcolor{color0!80}{0.14} & \textcolor{slow}{\sout{0.13}}     & 1.38                              & \textcolor{slow}{\textit{0.04}}   \\
			   & kn05  & \textcolor{color0}{\textbf{0.09}} & \textcolor{color0!80}{0.16} & \textcolor{slow}{\sout{0.14}}     & 0.18                              & \textcolor{slow}{\textit{0.05}}   \\
			   & nd01  & \textcolor{color0}{\textbf{0.07}} & 0.12                        & \textcolor{color0!80}{0.12}       & \textcolor{slow}{\sout{25.18}}    & \textcolor{slow}{\textit{0.06}}   \\
			   & nd02  & \textcolor{color0}{\textbf{0.05}} & 0.08                        & \textcolor{color0!80}{0.08}       & 0.15                              & \textcolor{slow}{\textit{0.04}}   \\
			   & nd10  & \textcolor{color0}{\textbf{0.06}} & \textcolor{color0!80}{0.10} & 0.10                              & \textcolor{slow}{\sout{39.62}}    & \textcolor{slow}{\textit{0.06}}   \\
			   & nn04  & \textcolor{color0}{\textbf{0.05}} & 0.09                        & \textcolor{color0!80}{0.08}       & \textcolor{slow}{\sout{13.18}}    & \textcolor{slow}{\textit{0.05}}   \\
			   & nn08  & \textcolor{color0}{\textbf{0.06}} & \textcolor{color0!80}{0.12} & 0.12                              & \textcolor{slow}{\sout{185.19}}   & \textcolor{slow}{\textit{0.05}}   \\
			   & nn13  & \textcolor{color0}{\textbf{0.09}} & \textcolor{color0!80}{0.15} & 0.24                              & \textcolor{slow}{\sout{187.08}}   & \textcolor{slow}{\textit{0.10}}   \\
			   & td02  & \textcolor{color0}{\textbf{0.09}} & 0.14                        & \textcolor{color0!80}{0.12}       & 1.32                              & \textcolor{slow}{\textit{0.04}}   \\
			   & td03  & \textcolor{color0}{\textbf{0.09}} & \textcolor{color0!80}{0.14} & \textcolor{slow}{\sout{0.11}}     & 0.37                              & \textcolor{slow}{\textit{0.04}}   \\
			   & td04  & \textcolor{color0}{\textbf{0.12}} & \textcolor{color0!80}{0.19} & \textcolor{slow}{\sout{0.14}}     & 2.95                              & \textcolor{slow}{\textit{0.05}}   \\
			   & tn07  & \textcolor{color0}{\textbf{0.11}} & \textcolor{color0!80}{0.17} & \textcolor{slow}{\sout{0.14}}     & 0.69                              & \textcolor{slow}{\textit{0.05}}   \\
			   & tn08  & \textcolor{color0}{\textbf{0.10}} & \textcolor{color0!80}{0.15} & \textcolor{slow}{\sout{0.12}}     & 1.13                              & \textcolor{slow}{\textit{0.05}}   \\
			   & tn09  & \textcolor{color0}{\textbf{0.11}} & \textcolor{color0!80}{0.17} & \textcolor{slow}{\sout{0.13}}     & 1.90                              & \textcolor{slow}{\textit{0.04}}   \\
			\midrule
			\multirow[c]{8}{*}{\rotatebox[origin=c]{90}{CU}}
			   & klr1  & \textcolor{color0}{\textbf{0.03}} & 0.04                        & \textcolor{color0!80}{0.04}       & 20.86                             & \textcolor{slow}{\textit{0.04}}   \\
			   & klr2  & \textcolor{color0}{\textbf{0.03}} & 0.04                        & \textcolor{color0!80}{0.03}       & \textcolor{slow}{\sout{~~~~}}     & \textcolor{slow}{\textit{0.04}}   \\
			   & klr3  & \textcolor{color0}{\textbf{0.03}} & 0.05                        & \textcolor{color0!80}{0.04}       & \textcolor{slow}{\sout{22.95}}    & \textcolor{slow}{\textit{0.04}}   \\
			   & klr4  & \textcolor{color0}{\textbf{0.03}} & 0.04                        & \textcolor{color0!80}{0.04}       & 5802.21                           & \textcolor{slow}{\textit{0.03}}   \\
			   & mcr1  & \textcolor{color0}{\textbf{0.05}} & 0.06                        & \textcolor{color0!80}{0.06}       & 29.15                             & \textcolor{slow}{\textit{0.06}}   \\
			   & mcr2  & \textcolor{color0}{\textbf{0.03}} & \textcolor{color0!80}{0.05} & 0.06                              & 3.20                              & \textcolor{slow}{\textit{0.03}}   \\
			   & mcr3  & \textcolor{color0}{\textbf{0.03}} & 0.06                        & 0.05                              & \textcolor{color0!80}{0.05}       & \textcolor{slow}{\textit{0.04}}   \\
			   & mcr4  & \textcolor{color0}{\textbf{0.03}} & \textcolor{color0!80}{0.05} & \textcolor{slow}{\sout{0.04}}     & \textcolor{slow}{\sout{3.52}}     & \textcolor{slow}{\textit{0.04}}   \\
			\midrule
			\multirow[c]{7}{*}{\rotatebox[origin=c]{90}{BG}}
			   & 0500  & \textcolor{color0}{\textbf{0.05}} & 0.09                        & 0.07                              & \textcolor{color0!80}{0.05}       & 0.05                              \\
			   & 0501  & \textcolor{color0!80}{0.05}       & 0.07                        & 0.07                              & \textcolor{color0}{\textbf{0.04}} & 0.05                              \\
			   & 0507  & \textcolor{color0}{\textbf{0.05}} & 0.08                        & 0.11                              & 0.06                              & \textcolor{color0!80}{0.05}       \\
			   & 0601  & \textcolor{color0!80}{0.05}       & 0.08                        & 0.07                              & 0.05                              & \textcolor{color0}{\textbf{0.05}} \\
			   & 0803  & 0.06                              & 0.09                        & 0.09                              & \textcolor{color0!80}{0.06}       & \textcolor{color0}{\textbf{0.05}} \\
			   & 1800  & \textcolor{color0!80}{0.05}       & 0.07                        & 0.07                              & \textcolor{color0}{\textbf{0.05}} & \textcolor{slow}{\textit{0.06}}   \\
			   & 1813  & 0.06                              & 0.11                        & 0.11                              & \textcolor{color0}{\textbf{0.05}} & \textcolor{color0!80}{0.05}       \\

			\bottomrule
		\end{tabular}
	\end{minipage}%
	\begin{minipage}{.5\linewidth}
		\centering
		\begin{tabular}{l|c||ccccc}
			\toprule
			\multicolumn{7}{c}{\rte{30}}                                                                                                                                                                   \\
			\toprule
			D. & Seq.  & \acs{ours}                        & KISS                              & GenZ                              & MAD                               & CT                                \\
			\midrule
			\multirow[c]{3}{*}{\rotatebox[origin=c]{90}{N20}}
			   & le    & \textcolor{color0}{\textbf{0.44}} & 0.46                              & 0.46                              & \textcolor{color0!80}{0.45}       & 0.64                              \\
			   & pm    & \textcolor{color0}{\textbf{0.54}} & \textcolor{color0!80}{0.57}       & \textcolor{slow}{\sout{1.95}}     & 1.01                              & \textcolor{slow}{\textit{29.16}}  \\
			   & se    & 0.38                              & 0.39                              & \textcolor{color0!80}{0.37}       & \textcolor{color0}{\textbf{0.37}} & \textcolor{slow}{\textit{0.55}}   \\
			\midrule
			\multirow[c]{9}{*}{\rotatebox[origin=c]{90}{N21}}
			   & c     & 0.53                              & 0.66                              & \textcolor{color0!80}{0.51}       & 0.53                              & \textcolor{color0}{\textbf{0.37}} \\
			   & me    & \textcolor{color0}{\textbf{0.14}} & \textcolor{color0!80}{0.14}       & 0.14                              & 0.24                              & 0.32                              \\
			   & mh    & \textcolor{color0!80}{1.12}       & 1.73                              & 14.58                             & 21.64                             & \textcolor{color0}{\textbf{0.51}} \\
			   & mm    & \textcolor{color0}{\textbf{0.21}} & 0.23                              & \textcolor{color0!80}{0.22}       & 18.34                             & 2.77                              \\
			   & p     & \textcolor{color0!80}{0.55}       & 0.66                              & 0.57                              & \textcolor{slow}{\sout{12.36}}    & \textcolor{color0}{\textbf{0.47}} \\
			   & qe    & \textcolor{color0!80}{0.49}       & 0.52                              & 0.54                              & 0.57                              & \textcolor{color0}{\textbf{0.37}} \\
			   & qh    & \textcolor{color0}{\textbf{0.93}} & 1.08                              & \textcolor{color0!80}{1.04}       & 6.07                              & 1.27                              \\
			   & qm    & \textcolor{slow}{\textit{0.79}}   & \textcolor{color0!80}{0.91}       & 0.96                              & \textcolor{color0}{\textbf{0.89}} & 17.43                             \\
			   & s     & 0.47                              & 6.72                              & \textcolor{color0!80}{0.39}       & \textcolor{color0}{\textbf{0.28}} & 132.51                            \\
			\midrule
			\multirow[c]{6}{*}{\rotatebox[origin=c]{90}{H22}}
			   & atug2 & \textcolor{color0}{\textbf{3.08}} & 14.69                             & 11.05                             & \textcolor{color0!80}{4.47}       & 33.38                             \\
			   & b2    & \textcolor{color0}{\textbf{0.78}} & 10.35                             & \textcolor{color0!80}{0.78}       & 1.50                              & 31.89                             \\
			   & clg2  & \textcolor{color0}{\textbf{0.89}} & 3.61                              & \textcolor{color0!80}{1.20}       & \textcolor{slow}{\sout{~~~~}}     & 13.25                             \\
			   & cul1  & \textcolor{color0}{\textbf{0.56}} & \textcolor{color0!80}{2.94}       & 3.82                              & 20.93                             & 20.54                             \\
			   & cul2  & \textcolor{color0}{\textbf{0.40}} & \textcolor{color0!80}{2.30}       & 9.37                              & 3.49                              & 3.34                              \\
			   & cul3  & \textcolor{color0}{\textbf{1.71}} & 36.27                             & \textcolor{slow}{\sout{22.42}}    & \textcolor{color0!80}{14.85}      & 22.90                             \\
			\midrule
			\multirow[c]{13}{*}{\rotatebox[origin=c]{90}{OS}}
			   & bl02  & 0.70                              & 0.83                              & 1.00                              & \textcolor{color0!80}{0.60}       & \textcolor{color0}{\textbf{0.37}} \\
			   & bp01  & \textcolor{color0}{\textbf{1.05}} & \textcolor{color0!80}{1.07}       & 2.47                              & 3.96                              & 10.59                             \\
			   & bp02  & 0.42                              & 0.39                              & \textcolor{color0!80}{0.39}       & \textcolor{color0}{\textbf{0.37}} & 1.23                              \\
			   & bp05  & \textcolor{color0!80}{1.30}       & 1.69                              & 3.30                              & \textcolor{color0}{\textbf{0.83}} & 19.41                             \\
			   & cc01  & 0.48                              & 0.60                              & 0.52                              & \textcolor{color0!80}{0.46}       & \textcolor{color0}{\textbf{0.24}} \\
			   & cc02  & \textcolor{color0!80}{0.46}       & 0.60                              & 1.33                              & 0.47                              & \textcolor{color0}{\textbf{0.41}} \\
			   & cc03  & \textcolor{color0!80}{0.46}       & 0.50                              & 0.50                              & 0.47                              & \textcolor{color0}{\textbf{0.30}} \\
			   & cc05  & \textcolor{color0}{\textbf{0.41}} & 0.54                              & 0.47                              & \textcolor{color0!80}{0.45}       & 1.02                              \\
			   & kc02  & 0.49                              & 0.82                              & \textcolor{color0!80}{0.49}       & \textcolor{color0}{\textbf{0.47}} & 8.50                              \\
			   & kc03  & \textcolor{color0}{\textbf{0.51}} & 4.70                              & 0.62                              & \textcolor{color0!80}{0.52}       & 3.45                              \\
			   & kc04  & \textcolor{color0!80}{0.55}       & 0.99                              & 1.16                              & \textcolor{color0}{\textbf{0.54}} & \textcolor{slow}{\textit{0.31}}   \\
			   & oq01  & \textcolor{color0!80}{0.45}       & 0.56                              & 0.48                              & 0.47                              & \textcolor{color0}{\textbf{0.23}} \\
			   & oq02  & \textcolor{color0!80}{0.48}       & 0.55                              & 0.54                              & \textcolor{color0}{\textbf{0.48}} & \textcolor{slow}{\textit{6.32}}   \\
			\midrule
			\multirow[c]{18}{*}{\rotatebox[origin=c]{90}{MC}}
			   & kd06  & \textcolor{color0}{\textbf{0.49}} & 0.51                              & \textcolor{color0!80}{0.51}       & 0.53                              & \textcolor{slow}{\textit{0.19}}   \\
			   & kd09  & \textcolor{color0}{\textbf{0.58}} & 0.62                              & \textcolor{slow}{\sout{0.62}}     & \textcolor{color0!80}{0.59}       & \textcolor{slow}{\textit{0.22}}   \\
			   & kd10  & \textcolor{color0}{\textbf{0.67}} & \textcolor{color0!80}{0.70}       & \textcolor{slow}{\sout{0.71}}     & 0.78                              & \textcolor{slow}{\textit{4.96}}   \\
			   & kn01  & \textcolor{color0}{\textbf{0.45}} & 0.47                              & \textcolor{color0!80}{0.46}       & 0.50                              & \textcolor{slow}{\textit{0.19}}   \\
			   & kn04  & \textcolor{color0}{\textbf{0.46}} & \textcolor{color0!80}{0.49}       & \textcolor{slow}{\sout{0.46}}     & 13.51                             & \textcolor{slow}{\textit{0.21}}   \\
			   & kn05  & \textcolor{color0}{\textbf{0.51}} & 0.57                              & \textcolor{slow}{\sout{0.48}}     & \textcolor{color0!80}{0.57}       & \textcolor{slow}{\textit{0.23}}   \\
			   & nd01  & \textcolor{color0}{\textbf{0.55}} & \textcolor{color0!80}{0.57}       & 0.59                              & \textcolor{slow}{\sout{446.02}}   & \textcolor{slow}{\textit{0.72}}   \\
			   & nd02  & 0.36                              & \textcolor{color0!80}{0.34}       & \textcolor{color0}{\textbf{0.34}} & 0.49                              & \textcolor{slow}{\textit{0.25}}   \\
			   & nd10  & \textcolor{color0}{\textbf{0.50}} & \textcolor{color0!80}{0.51}       & 0.54                              & \textcolor{slow}{\sout{575.34}}   & \textcolor{slow}{\textit{0.30}}   \\
			   & nn04  & \textcolor{color0}{\textbf{0.44}} & \textcolor{color0!80}{0.44}       & 0.45                              & \textcolor{slow}{\sout{146.22}}   & \textcolor{slow}{\textit{0.25}}   \\
			   & nn08  & \textcolor{color0}{\textbf{0.52}} & \textcolor{color0!80}{0.57}       & 0.68                              & \textcolor{slow}{\sout{2014.48}}  & \textcolor{slow}{\textit{0.25}}   \\
			   & nn13  & \textcolor{color0}{\textbf{0.56}} & \textcolor{color0!80}{0.59}       & 1.73                              & \textcolor{slow}{\sout{3164.78}}  & \textcolor{slow}{\textit{1.09}}   \\
			   & td02  & \textcolor{color0}{\textbf{0.47}} & 0.49                              & \textcolor{color0!80}{0.49}       & 3.03                              & \textcolor{slow}{\textit{0.17}}   \\
			   & td03  & \textcolor{color0}{\textbf{0.51}} & \textcolor{color0!80}{0.54}       & \textcolor{slow}{\sout{0.52}}     & 1.32                              & \textcolor{slow}{\textit{0.18}}   \\
			   & td04  & \textcolor{color0}{\textbf{0.54}} & \textcolor{color0!80}{0.58}       & \textcolor{slow}{\sout{0.56}}     & 23.74                             & \textcolor{slow}{\textit{0.20}}   \\
			   & tn07  & \textcolor{color0}{\textbf{0.58}} & \textcolor{color0!80}{0.62}       & \textcolor{slow}{\sout{0.60}}     & 2.60                              & \textcolor{slow}{\textit{0.21}}   \\
			   & tn08  & \textcolor{color0}{\textbf{0.51}} & \textcolor{color0!80}{0.54}       & \textcolor{slow}{\sout{0.54}}     & 10.61                             & \textcolor{slow}{\textit{0.18}}   \\
			   & tn09  & \textcolor{color0}{\textbf{0.49}} & \textcolor{color0!80}{0.51}       & \textcolor{slow}{\sout{0.51}}     & 19.59                             & \textcolor{slow}{\textit{0.20}}   \\
			\midrule
			\multirow[c]{8}{*}{\rotatebox[origin=c]{90}{CU}}
			   & klr1  & 0.30                              & \textcolor{color0!80}{0.25}       & \textcolor{color0}{\textbf{0.25}} & 23.31                             & \textcolor{slow}{\textit{0.28}}   \\
			   & klr2  & 0.16                              & \textcolor{color0}{\textbf{0.13}} & \textcolor{color0!80}{0.13}       & \textcolor{slow}{\sout{~~~~}}     & \textcolor{slow}{\textit{0.21}}   \\
			   & klr3  & 0.20                              & \textcolor{color0!80}{0.18}       & \textcolor{color0}{\textbf{0.17}} & \textcolor{slow}{\sout{37.74}}    & \textcolor{slow}{\textit{0.22}}   \\
			   & klr4  & 0.21                              & \textcolor{color0!80}{0.15}       & \textcolor{color0}{\textbf{0.15}} & 7087.17                           & \textcolor{slow}{\textit{0.20}}   \\
			   & mcr1  & 0.18                              & \textcolor{color0!80}{0.17}       & \textcolor{color0}{\textbf{0.17}} & 208.30                            & \textcolor{slow}{\textit{0.21}}   \\
			   & mcr2  & 0.18                              & \textcolor{color0!80}{0.16}       & \textcolor{color0}{\textbf{0.16}} & 4.39                              & \textcolor{slow}{\textit{0.20}}   \\
			   & mcr3  & \textcolor{color0!80}{0.21}       & 0.22                              & \textcolor{color0}{\textbf{0.21}} & 0.21                              & \textcolor{slow}{\textit{0.22}}   \\
			   & mcr4  & \textcolor{color0!80}{0.18}       & \textcolor{color0}{\textbf{0.16}} & \textcolor{slow}{\sout{0.17}}     & \textcolor{slow}{\sout{4.78}}     & \textcolor{slow}{\textit{0.22}}   \\
			\midrule
			\multirow[c]{7}{*}{\rotatebox[origin=c]{90}{BG}}
			   & 0500  & 0.81                              & 1.33                              & 1.48                              & \textcolor{color0!80}{0.79}       & \textcolor{color0}{\textbf{0.75}} \\
			   & 0501  & \textcolor{color0!80}{0.78}       & 1.42                              & 1.49                              & 0.80                              & \textcolor{color0}{\textbf{0.77}} \\
			   & 0507  & \textcolor{color0!80}{0.82}       & 1.13                              & 1.55                              & 0.82                              & \textcolor{color0}{\textbf{0.76}} \\
			   & 0601  & 0.82                              & 1.10                              & 1.17                              & \textcolor{color0!80}{0.78}       & \textcolor{color0}{\textbf{0.74}} \\
			   & 0803  & 0.86                              & 1.03                              & 1.25                              & \textcolor{color0!80}{0.84}       & \textcolor{color0}{\textbf{0.74}} \\
			   & 1800  & \textcolor{color0}{\textbf{0.80}} & 1.06                              & 1.09                              & \textcolor{color0!80}{0.82}       & \textcolor{slow}{\textit{0.82}}   \\
			   & 1813  & 0.83                              & 1.09                              & 1.44                              & \textcolor{color0!80}{0.81}       & \textcolor{color0}{\textbf{0.77}} \\

			\bottomrule
		\end{tabular}
	\end{minipage}%
\end{table*}

\subsection{Ablations}
We perform an ablation to show the impact of optimizing a window of poses rather than just filtering the current pose. The rest of the pipeline, including feature extraction, map aggregation, etc., all remained the same. The summarized results can be seen inf Table~\ref{tab:ablation_smoothing}. As expected, smoothing requires more compute, but does improve performance across all datasets, particularly for the longer windowed \rte{30} metric. Interestingly, it had a more minimal impact on trajectory smoothness; this could be due to using the pose estimates immediately rather than upon marginalization as mentioned previously.

Additionally, we perform an ablation comparing \ac{ours} with only planar features to the version with both planar and point features. The results are summarized in Table~\ref{tab:ablation_feature}. Generally, the addition of points features reduced drift amounts in the longer windowed \ac{wrte}, particularly for datasets with unstructured environments such as \acf{mc} and \acf{os}, while having a limited impact on trajectory smoothness. The exception to this is the extremely structured datasets \acf{h22} and \acf{n21}, where point features were less impactful.

\subsection{Comparison with Prior Works}
We additionally run on 64 trajectories across the seven datasets listed in Table~\ref{table:datasets}. The main results are summarized in Table~\ref{tab:baseline_main}

\begin{figure}[t]
    \centering
    \includegraphics[width=0.48\textwidth]{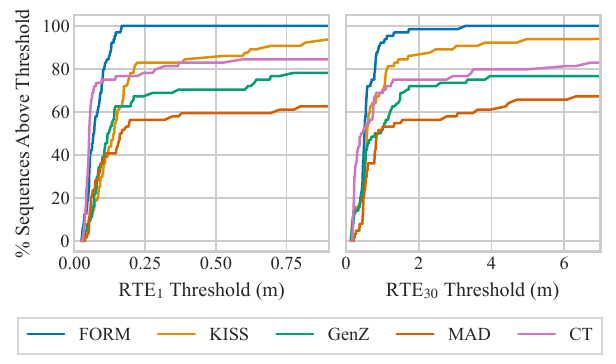}
    \caption{Percentage of sequences for each pipeline that falls below the given \rte{1} and \rte{30} thresholds. Note this does not take into account sequences with sub-real-time performance. \ac{ours} provides smooth trajectories as shown by \rte{1} below \SI{20}{\centi\meter} on all sequences, and is the sole pipeline to have successfully completed all sequences.}\label{fig:curve}
    \vspace{-1.0em}
\end{figure}

Representing odometry drift, \rte{30} results can be seen in the right of Table~\ref{tab:baseline_main}. \ac{ours} is competitive in this category and is generally only a few centimeters away from the best method. Importantly, \ac{ours} never has a significant failure case and performs well across all environments with identical parameters. The is well-summarized in Fig.~\ref{fig:curve}, which shows success rates for different \rte{30} thresholds. In this figure we can see that \ac{ours} clearly has the largest area under the curve and is the only pipeline to successfully complete all trajectories at a success threshold of \SI{3.08}{\meter}.

Importantly, \ac{ours} is easily the leader in \rte{1}, often by a significant margin. This is well summarized in the left of Fig.~\ref{fig:curve} and Table~\ref{tab:baseline_main}. For some sequences, such as bp01 in \acl{os}, the \rte{1} of \ac{ours} is a mere fraction of that of KISS-ICP and GenZ-ICP despite having similar \rte{30}. This is of crucial importance, as these trajectories will often be used by downstream tasks such as planning, control, or even initialization of the next step of the pipeline. Jittery trajectories, as seen in Fig.~\ref{fig:window}, can cause inaccurate estimates of velocity and position, and significantly degrade robustness of the overall systems.

Finally, we compare the rate of computation to ensure real-time performance. These rates are averaged across datasets, while excluding trajectories with \rte{30} greater than ten meters, as often those sequences run at irregular rates due to failure. The results are summarized in Table~\ref{tab:baseline_speed}. Both KISS-ICP and GenZ-ICP are by far the fastest, which is to be expected since they are simple \ac{icp} methods. On the other side, MAD-ICP ran almost exactly at real-time due to its ``real-time'' mode. CT-ICP struggled to be real-time in trajectories it finished successfully, most notably \acl{mc} and \acl{cu}. On other datasets, it often failed and caused the pipeline to run super-real-time. \ac{ours} falls somewhere in between these two camps; it does require more compute time than the simpler \ac{icp} methods, but runs comfortably real-time, particularly for 64-beam lidars datasets.

As a note on prior works, GenZ-ICP failed to complete a few trajectories due to accumulating a map of over a million points. We considered tuning the voxel size to fix this, but didn't want to degrade performance on other sequences so default parameters from the repository were used. Additionally, when CT-ICP works, it performs very well. Unfortunately, it either diverged or was sub-real-time a significant amount of the time. We suspect this could be fix  on a per dataset scenario via parameter tuning; however, we feel it is an important benchmark that a \ac{lo} pipeline works across datasets without additional tuning.

\section{Conclusion}\label{sec:finale}

In this work, we have proposed \ac{ours}, a novel \ac{lo} method that smooths over a past window of poses, iteratively performs map corrections as poses are smoothed, and still provides real-time performance due to matching against a single-map. We provided ablations to validate our design choices, specifically that of smoothing and feature selection, to show they increase robustness and accuracy. Additionally, we empirically show across over 60 sequences and seven datasets that \ac{ours} provides competitive long-term drift accuracy, robust performance, and still maintains smooth trajectory estimates across all datasets.

We believe this line of work holds the potential for further improvements. The feature extraction was mostly chosen as a safe, well-trusted technique, and we believe there is still significant work that can be done to improve performance and make it functional with solid-state \ac{lidar}s. It should also be noted that sensor fusion is straightforward with \ac{ours} due to its underlying factor graph representation, an important feature as \ac{lidar} sensor fusion is currently an active area of research. Due to this, we plan to pursue the trivial fusion in a factor graph with other sensors such as IMUs, wheel encoders, and others.

\IEEEtriggeratref{10} 

\bibliographystyle{ieeetr}
\footnotesize
\bibliography{ref}

\end{document}